\documentclass[journal]{IEEEtran}

\usepackage{cite}

\ifCLASSINFOpdf
   \usepackage[pdftex]{graphicx}
   \graphicspath{{./figures/}}
   \DeclareGraphicsExtensions{.pdf,.jpeg,.png}
\else
   \graphicspath{{./figures/}}
\fi

\usepackage{amsmath}
\usepackage{amssymb}
\usepackage{algorithmic}
\usepackage[ruled,vlined]{algorithm2e}
\usepackage{array}
\usepackage{multirow}
\usepackage{url}

\begin{document}

\title{Unsupervised Domain Adaptive Person Re-ID with Local-enhance and Prototype Dictionary Learning}

\author{Haopeng Hou,
        }
\thanks{.} 
\thanks{.}

\markboth{Journal of \LaTeX\ Class Files,~Vol., No., XX XXXX}%
{Shell \MakeLowercase{\textit{et al.}}: Bare Demo of IEEEtran.cls for IEEE Journals}

\maketitle

\begin{abstract}
  The unsupervised domain adaptive person re-identification (re-ID) task has been a challenge because, unlike the general domain adaptive tasks, there is no overlap between the classes of source and target domain data in the person re-ID, which leads to a significant domain gap. State-of-the-art unsupervised re-ID methods train the neural networks using a memory-based contrastive loss. However, performing contrastive learning by treating each unlabeled instance as a class will lead to the problem of class collision, and the updating intensity is inconsistent due to the difference in the number of instances of different categories when updating in the memory bank. To address such problems, we propose Prototype Dictionary Learning for person re-ID which is able to utilize both source domain data and target domain data by one training stage while avoiding the problem of class collision and the problem of updating intensity inconsistency by cluster-level prototype dictionary learning. In order to reduce the interference of domain gap on the model, we propose a local-enhance module to improve the domain adaptation of the model without increasing the number of model parameters. Our experiments on two large datasets demonstrate the effectiveness of the prototype dictionary learning. 71.5\% mAP is achieved in the Market-to-Duke task, which is a 2.3\% improvement compared to the state-of-the-art unsupervised domain adaptive re-ID methods. It achieves 83.9\% mAP in the Duke-to-Market task, which improves by 4.4\% compared to the state-of-the-art unsupervised adaptive re-ID methods.
\end{abstract}

\begin{IEEEkeywords}
  Person re-identification, Unsupervised domain adaptation, Prototype dictionary learning, Local-enhance.
\end{IEEEkeywords}

\IEEEpeerreviewmaketitle

\section{Introduction}

\IEEEPARstart{P}{erson} re-identification (re-ID) aims to identify whether photos captured by different cameras are from the same person. Person re-ID works based on supervised learning \cite{{chang2018multi},{chen2018person},{7289409},{8315486},{zheng2019joint},{8693885},{wang2018learning},{zhou2019omni}} have achieved high performance. However, the labeled data set necessary for supervised methods is very expensive to collect. The unsupervised learning approaches \cite{lin2019a} \cite{yu2017cross} can use almost zero-cost unlabeled data, but due to the lack of supervision with real labels, the results are not satisfactory. Also, the unsupervised approaches have a demand for high-level hardware during training which is currently not conducive to practical applications.

To tackle this problem, the unsupervised domain adaptation (UDA) methods were proposed which transfer the knowledge learned from the source domain (labeled images) to the target domain (unlabeled images). The pseudo-label-based methods \cite{{zhai2020ad},{zhang2019self},{fu2019self},{ge2020mutual}} of of UDA for person re-ID have proven to be effective and maintains the best performance so far. Existing UDA methods on person re-ID uses two stages to train the model: first train a pre-trained model on the source domain with labeled data, and then fine-tune it on the target domain with unlabeled data. However, the two-stage strategy training is not sufficient for mining labeled data and the performance is not satisfactory when the difference between the two domains is significant. To address this problem, Ge \emph{et al}. proposes the self-paced contrastive learning framework (SpCL) \cite{ge2020self}, which combines the pre-training stage and fine-tuning stage to train both labeled and unlabeled data. Moreover, in order to reasonably mine all available information, contrastive learning was introduced. SpCL stores the feature vectors in the hybrid memory and calculates the loss using the unified contrastive loss, then updates the network parameters by back-propagating and the feature vectors in the hybrid memory by momentum-updating. Although SpCL achieves considerable improvement, we argue that there are still three drawbacks. First, instance-level features are often used for comparison in traditional contrastive learning, a large number of negative instances are needed to avoid model collapse \cite{{he2020momentum},{chen2020simple}}. However massive negative instances inevitably create the class collision problem \cite{saunshi2019}. Because sampling a large number of negative instances inevitably generates negative pairs with similar semantics, which should be close to each other in the embedding space but be separated by the instance-level contrastive learning as the instance-level contrastive learning treats any two samples as a negative pair as long as they are different instances \cite{li2020prototypical}. Second, the updating frequencies of feature prototypes in the memory bank are inconsistent, due to the inconsistency in the number of samples under different classes of persons which is more serious in the early training period when clustering unlabeled data. The prototypes corresponding to the clusters containing more samples updates more frequently. Third, due to the presence of the domain gap, simultaneous training with the source domain data and the target domain data may interfere with model convergence. By reducing the domain gap between the source and target domains through style normalization, the fine-tuning of the source domain model on the target domain can be achieved more easily, and the generation and interference of the wrong pseudo-labels can also be mitigated. Huang \emph{et al}. \cite{huang2017arbitrary} proposed that the style of an image is the statistical information, such as mean and variance, of each feature channel of the feature map. Thus the domain gap between the two domains can be changed by changing the mean and variance of the features of the two domains. 

To overcome these three problems, we propose Prototype Dictionary Learning (PDL). We propose a prototype memory to store feature prototypes of both classes in the source domain and clusters in the target domain. The so-called prototype is a representative embedding of a set of semantically similar instances. We use the class centroids of the true classes of the source domain as the class prototypes of the source domain data, and the cluster centroids of the target domain clusters as the class prototypes of the target domain data. It replaces the instance-level feature with the class/cluster-level feature to avoid the problem of class collision. Meanwhile, we propose a corresponding loss function named PDL loss to guide model optimization during the training process with both source domain and target domain. To alleviate the amplification of inconsistency in the frequency of updating the memory dictionary during training, PDL reinitializes the memory dictionary before each epoch. Lastly, we propose a local-enhance module to change the value of a random part of the feature map which improves the adaptability of the model to different domain styles during training, resulting in increased domain generalization of the model without increasing parameters.

In summary, the main contributions of this paper are :
\begin{itemize}
 \item We propose the Prototype Dictionary Learning framework for person re-ID. It simplifies the training process, avoids category collision, and alleviates the prototype update inconsistency problem.
  \item We propose the PDL loss, which can realize class/cluster-level contrastive learning in the process of training with both labeled and unlabeled data. 
  \item We propose a local-enhance module. It can increases domain generalization of the model without increasing parameters. 
  \item We evaluate our method on Market-1501 and DukeMTMC-reID. Experiments show that our method achieves competitive results in UDA person re-ID. 
\end{itemize}

\section{Related Work}
\subsection{Contrastive Learning}
Since Raia Hadsell \emph{et al}. \cite{hadsell2006dimensionality} proposed the approach of contrastive learning in 2006 which is a supervised contrastive learning method, contrastive learning has received a lot of attention and research. Broad contrastive learning includes supervised contrastive learning \cite{{varior2016gated},{hermans2017defense}} and unsupervised contrastive learning, while narrow contrastive learning refers to self-supervised contrastive learning. The contrastive learning mentioned in this paper refers to narrow contrastive learning. Contrastive learning can be broadly divided into two categories: negative-example-based contrastive learning methods \cite{{chen2020simple},{he2020momentum}}and clustering-based contrastive learning methods \cite{{caron2020unsupervised},{li2020prototypical}}. The process of negative-example-based contrastive learning can be roughly divided into image enhancement, feature encoding, and loss calculation. Image enhancement is a common method for contrastive learning to construct similar and dissimilar samples by itself. In order to prevent model collapse, SimCLR \cite{chen2020simple} uses a large batch size, MOCO \cite{he2020momentum} introduced memory bank to avoid the use of large batch size while improving performance. The process of the cluster-based contrastive learning methods is roughly the same as that of the negative-example-based contrastive learning methods. Instead of using computationally intensive pairwise comparisons, cluster-based contrastive learning obtains feature prototypes from the clustering assignment and contrasts features prototype and sample features. The PCL \cite{li2020prototypical} uses clustering directly to obtain similar and dissimilar sample sets.

\subsection{Unsupervised Domain Adaptation for Person Re-ID}
To solve the problems of the high cost of labeled data acquisition and poor generalization of person re-identification systems, unsupervised domain adaptive (UDA) methods are introduced. Unsupervised domain adaptive person re-identification can be divided into two main categories: the style transfer methods and the pseudo label estimation methods. Both SPGAN \cite{deng2018image} proposed by Deng \emph{et al}. and PTGAN \cite{wei2018person} proposed by Wei \emph{et al}. utilize the generative adversarial network (GAN) to transfer the style of the source domain image to the style of the target domain while maintaining the original identity label. The style-transformed images and their identity labels are then used to fine-tune the model in a supervised manner. HHL \cite{zhong2018generali} proposed by Zhong \emph{et al}. learned camera-invariant features by camera-style shifting images. However, the performance of these methods deeply depends on the quality of image generation, and they do not explore the complex relationships between different samples in the target domain. Compared to the former, pseudo label estimation is more effective in capturing the target domain distribution. Pseudo-label estimation methods can be classified into clustering-based methods and instance-feature-similarity-based methods. Yang Fu \emph{et al}. \cite{fu2019self} construct the feature coordinate space with the help of an auxiliary dataset, then embed unlabeled data in the feature space to obtain soft multi-labels of unlabeled data. SSG \cite{fu2019self} uses the potential similarity of the whole body and local parts of unlabeled samples to automatically build multiple clusters from different perspectives. Yixiao Ge \emph{et al.} \cite{{ge2020mutual}} use soft pseudo labels obtained from the CNN and hard pseudo labels obtained from the clustering method for better feature learning. The pseudo label estimation methods based on clustering have proved to be more effective and maintain the most advanced accuracy \cite{fu2019self,ge2020mutual}.

\subsection{Memory Bank Mechanism}
The end-to-end update by back-propagation \cite{{Aar2018Representation},{chen2020simple},{Cheng2019LocalAF}} is a natural mechanism for contrastive learning. However, the approach requires a large mini-batch size to avoid model collapse. And this requires a large GPU memory size\cite{he2020momentum}. Xiao \emph{et al}. first used the memory bank mechanism in the task of person search using Online Instance Matching (OIM) loss to solve the training difficulties caused by sparse and unbalanced labeled data samples. Wu \emph{et al}. introduces a memory bank mechanism for unsupervised feature learning through non-parametric instance discriminations \cite{ZhirongWu2018UnsupervisedFL}. Instance-level feature vectors are stored in a memory bank. The softmax distribution is approximated by noise-contrast estimation (NCE) to achieve parametric-free multi-classification. He et al. propose the Momentum Contrast (MOCO) method which maintains the key representations in a queue, reducing the capacity of the memory bank while achieving a large capacity dictionary. Yixiao Ge \emph{et al.} propose the self-paced contrastive learning (SpCL) \cite{ge2020self} method that stores both instance-level features and class-level features in a memory bank called hybrid memory which enables mining of data distribution and hard samples.

\section{Methodology}
To simplify the training process of the UDA method, avoid the class collision \cite{saunshi2019} problem caused by instance-level contrastive learning, and alleviate the inconsistency when the memory dictionary is updated, we propose a prototype dictionary learning (PDL) framework. PDL only uses category-level supervision, including source-domain class-level supervision and target-domain cluster-level supervision. A local-enhance module is introduced to reduce the interference of domain gap when using two domain data for training at the same time. The content of this section is organized as follows: the overview of the proposed framework is introduced in Sec. \ref{sec:framework}, then we illustrate PDL loss in Sec. \ref{sec:PDL_Loss}. Lastly, we introduce the Local-enhance module in Sec. \ref{sec:local_enhance}. 
\subsection{Framework Overview}\label{sec:framework}
The framework consists of two main parts: a CNN-based feature encoder and a novel prototype dictionary which is the base of the category-level contrastive learning. Ge \emph{et al.} proposes the SpCL framework which jointly trains the encoder with all the source-domain class-level, target-domain cluster-level, and target-domain un-clustered instance-level supervisions \cite{ge2020self}. We argue that the instance-level supervision is harmful to robust feature learning as the contrastive learning treats each instance as a new class, leading to an increase in the distance of two instances of the same class which, to a certain extent, destroys the certainty of clustering. The training process of PDL is simple and can be roughly divided into two steps: the initialization of the prototype dictionary (Section \ref{sec:prototype_dictionary_1}) and the update of the prototype dictionary (Section \ref{sec:prototype_dictionary_2}). After the prototype dictionary is initialized, a mini-batch containing source and target domain data is input to the network, and the feature vector is extracted by the encoder, and then compared to the feature prototype in the prototype dictionary as query, and finally, the prototype dictionary and encoder are updated.  The details of the algorithm are presented in  Algorithm \ref{algorithm}. 
\begin{algorithm}[!htb]
  \caption{Prototype Dictionary Learning.}
  \label{algorithm}
  \textbf{Input:} Labeled source domain dataset $D_s$;
    
  \ \ \ \ \ \ \ \ \ \ \ Unlabeled target domain dataset $D_t$; 
    
  \ \ \ \ \ \ \ \ \ \ \ Maximun iteration $I_{max}$ and maximun epoch $Epochs$;
    
  \textbf{Parameters:} Ensembling momentum $m$, weighting factors $\lambda$\;
  \textbf{Output:} $Encoder$\;
  \textbf{Initialization:} Initialize $Encoder$ with ImageNet pre-trained weights \;
  
  \For {i=1: Epochs}{
    1:Extract features of data on $D_s$ and $D_t$ by $Encoder$ \;
    2:Generate pseudo labels for $D_t$ by DBSCAN \;
    3:Generate prototypes $p_s$ and $p_t$ by the ground-truth labels and pseudo labels respectively;
    4:Initialize $Prototype$ $Dictionary$ with $p_s$ and $p_t$\;
      \For{j=1:$I_{max}$}{
        1:Extract features on the mini-batch \;
        2:Update $Prototype$ $Dictionary$ according tr Eq. \ref{momentum_update} \;
        3:Update $Encoder$ according to Eq. \ref{pdlloss} \;
      }
  }
\end{algorithm}

\subsubsection{Prototype Dictinoary Initialization}\label{sec:prototype_dictionary_1}
\begin{figure*}[t]
  \centering
  \includegraphics[width=350pt]{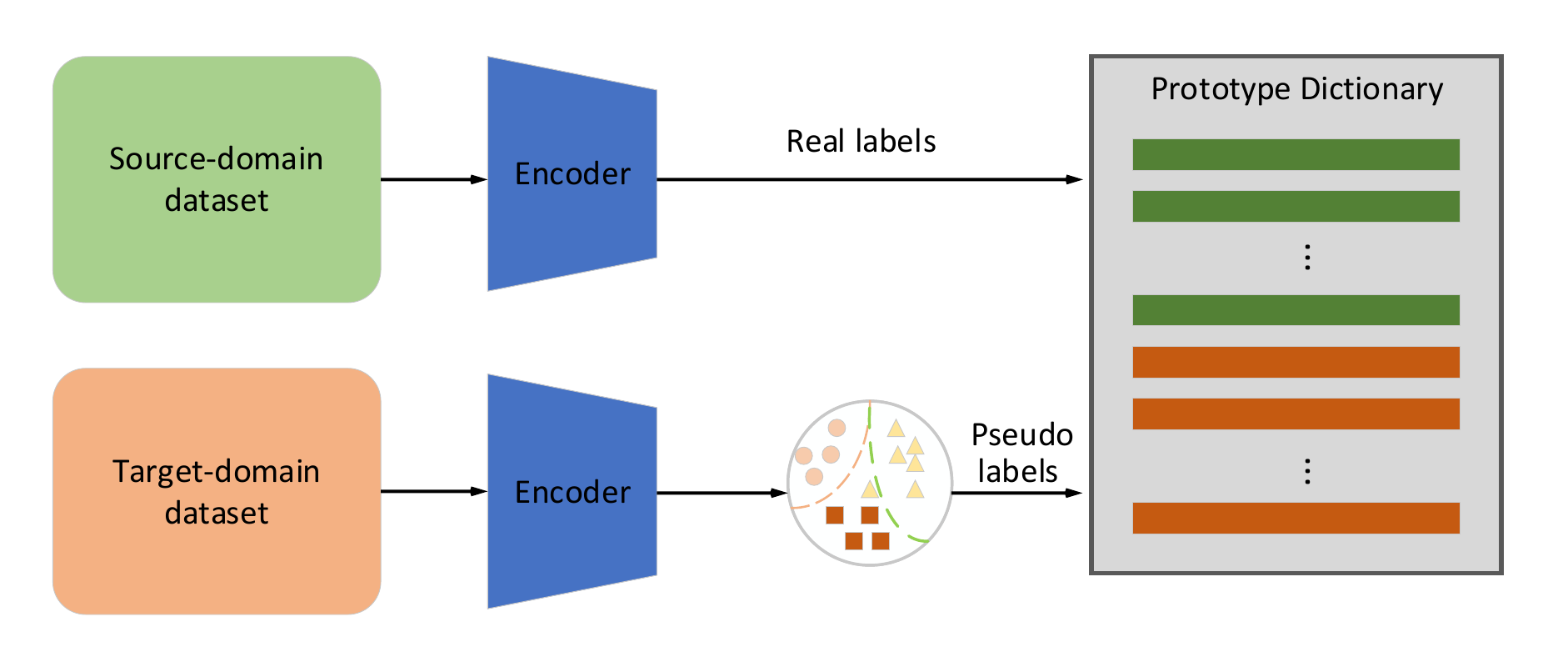}
  \centering
  \caption{Prototype dictionary initialization pipline. In order to facilitate understanding, we have drawn two encoders. In fact, the encoder respectively performs forward computation to extract the features of the date in source domain and target domain.}
  \label{pdl_1}
\end{figure*}

The prototype dictionary is initialized before each epoch. As illustrated in Fig. \ref{pdl_1}, the source domain data ${x^s} \in  \mathbb{X}^{s}$ and target domain data ${x^t} \in\mathbb{X}^{t}$ are input into the encoder $f_\theta$ at the same time. We use $f_\theta(x^s)$ to denote the feature vectors of the source domain data extracted by the encoder $f_\theta$ , and $f_\theta(x^t)$ to denote the feature vector corresponding to the target domain images. Then a clustering algorithm is applied to group the unlabeled features $f_\theta(x^t)$ into clusters. Specifically, we use DBSCAN \cite{MartinEster1996ADA} as the clustering algorithm which divides density-connected sample features into clusters without artificially designating the number of clusters. According to the ground-truth labels $\{l^s_1,\cdots,l^s_{n^s}\}$ of the source domain data and the pseudo-labels $\{l^t_1,\cdots,l^t_{n^t}\}$ of the target domain data generated by the clustering algorithm, the PDL calculates the centroids of the feature vectors in each category. We consider the centroids of each category as the feature prototypes of each category which is calculated by the following formula:
\begin{align}\label{prototype}
  {p}_{k}=\frac{1}{{N}_{k}} \sum_{{v}_{i} \in {\mathbb{X}}_{k}} {v}_{i},
\end{align}
where $\mathbb{X}_{k}$ denotes the feature vectors set containing all the feature vectors with the same label $l_k$ and the ${N}_{k}$ denotes the number of feature vectors belonging to the set.

Then the prototype dictinoary is initialized by caching source-domain prototypes $\{p^s_1,\cdots,p^s_{n^s}\}$ and target-domain prototypes $\{p^t_1,\cdots,p^t_{n^t}\}$. The capacity of prototype dictinoary is $n^s+n^t$. 

\subsubsection{Prototype Dictinoary Updating}\label{sec:prototype_dictionary_2}

\begin{figure*}[t]
  \centering
  \includegraphics[width=350pt]{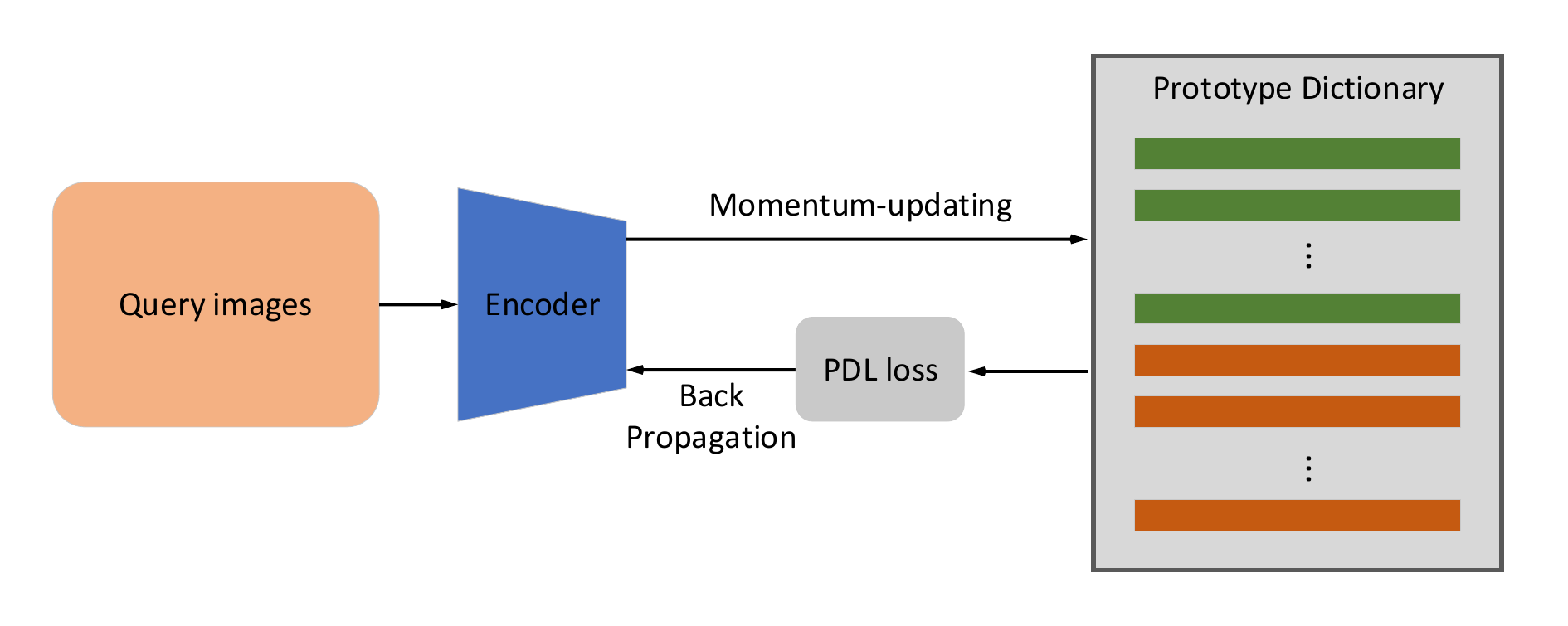}
  \centering
  \caption{Prototype dictionary updating pipline. The query images contains data of both source domain and targe domain.}
  \label{pdl_2}
\end{figure*}

The prototype dictionary is updated after every iteration as shown in Figure \ref{pdl_2}. Given a query image feature $f_\theta(x)$, the PDL computes the PDL loss by comparing the $f_\theta(x)$ with all the prototypes in the prototype dictionary. Then $f_\theta(x)$ is added to the corresponding prototype in a momentum-updating manner. 

The prototype can be calculated through the formula as follows: 
\begin{align}\label{momentum_update}
  p_k^{T}=m p_k^{T-1}+(1-m) f_{\theta}(x_i),
\end{align}
where $f_{\theta}(x_i)$ represents a query image feature belonging to the $k$-th category, $p_k$ represents the prototype of the $k$-th category, $T$ indicates the current iteration while $T-1$ indicates the previous iteration. $m$ is the ensembling momentum to be within the range $[0, 1)$. 

\subsection{PDL loss}\label{sec:PDL_Loss}
Given the source-domain training set $\mathbb{X}^{s}=\{x_1,\cdots,x_{n^s}\}$ and the target-domain training set $\mathbb{X}^{t}=\{x_{n^s+1},\cdots,x_{n^s+n^t}\}$, the encoder $f_\theta$ maps $\mathbb{X}^{s}$ and $\mathbb{X}^{t}$ to $V=\{v_1,\cdots,v_{N}\}$ where $N=n^s+n^t$. The training goal is to find the optimal $f_\theta$ such that $v$ best represents $x$. Traditional contrastive learning, such as InfoNCE loss \cite{CPC} shown in formula \ref{InfoNCE} used in SimCLR \cite{chen2020simple}, uses instance-level feature comparison, and does not explore the distribution characteristics of the data set itself. More importantly, instance-level comparison inherently treats each instance as a separate category, inevitably pushing the pair of samples that should be closer together. Saunshi \emph{et al.} \cite{saunshi2019} has proven that the phenomenon is harmful to representation learning. 
\begin{align}\label{InfoNCE}
  \mathcal{L}_{InfoNCE}=-\sum_{i=1}^{N}log\frac{exp (S(v_{i}, v_{i}^{+}) / \tau) }{ \sum_{j=0}^{K} exp (S(v_{i}, v_{j}) / \tau)}.
\end{align}
In the formula \ref{InfoNCE}, $N$ represents the number of the training data, $S(\cdot)$ means similarity calculation function. $\tau$ is the temperature. There are $K+1$ features involved in the comparison, of which 1 is a known positive sample of $v_i$, such as itself, and the remaining $k$ are regarded as negative samples. It is easy to find that this mechanism can only work when the sample size involved in the comparison is large. That is why SimCLR needs a large batch size and a lot of training epochs. 

Instance-level comparisons are easily disturbed by misjudged negative samples, and the clustering model naturally has a good tolerance for misjudged negative samples. We propose the PDL loss which is a kind of InfoNCE-like contrastive loss. The difference is that we use category-level comparisons instead of instance-level comparisons. The formula is as follows:
\begin{align}\label{pdlloss}
  \mathcal{L}_{PDL}=-\sum_{i=1}^{N} log \frac{exp (S(v_{i}, p_{i}^{+} / \tau) } {\sum_{j=1}^{n^s} exp (S(v_{i}, p_{j}) / \tau)+\sum_{r=1}^{n^t} exp (S(v_{i}, p_{r}) / \tau)},
\end{align}
where $p_i^+$ is the prototype of the category containing feature vector $v_i$, and $p_j$ and $p_r$ are prototypes of source-domain categories and target-domain categories respectively. In this way, we convert instance-level comparisons into category-level comparisons. The prototypes of the target domain come from the clustering algorithm, which enables PDL to strengthen the mining of the data distribution of the target domain. At the same time, the source-domain prototypes come from their own ground-truth labels, which can help the model to obtain the representative target domain prototypes. Only $n^s+n^t$ feature prototypes need to be stored and only $n^s+n^t$ feature prototypes are involved in the comparison. So the training batch size does not need to be large. In fact, the results shown in Table \ref{comparison_1} are obtained at a batch size of 64.

\subsection{Local-enhance Module}\label{sec:local_enhance} 
In order to reduce the interference of the domain gap on the model convergence when training with both the source domain data and the target domain data at the same time, and enhance the model's adaptability to different domains, we propose the Local-enhance Module. X.Huang \emph{et al.} \cite{huang2017arbitrary} proposed that the artistic style of the image is the statistical information of each feature channel across the space of the feature map, such as the mean and variance. Our motivation is to change the mean and variance of the feature map by randomly adding disturbances to the feature map so that the model reduces the sensitivity to image style and improves the generalization of the model between different domains. 

Person re-ID tasks are affected by posture changes and occlusion problems, and the local features are quite different. Therefore, the person re-ID network will suppress the local features during training, but some useful information will also be discarded. Inspired by the DropBlock \cite{ghiasi2018dropblock} mechanism, the local-enhance module select a contiguous region of a feature map. The local-enhance module enhances a contiguous region of the feature maps in a mini-batch instead of dropping them. 

We use ResNet50 as the backbone network and insert the local-enhance module between layer3 and layer4. Given the output tensor $\mathcal{F} \in \mathbb{R}^{C\times H \times W}$ of layer3 with the shape of $C\times H \times W$, where $C$ is the number of channels, $H$ and $W$ are the height and width respectively. The width of the enhanced block is the same as the width $W$ of the feature map. The height of the enhanced block is set to one-tenth of the height $H$ of the tensor. Therefore, the randomness of the region is only reflected in the vertical dimension. The enhancement operation can be expressed as follows as the enhancing coefficient is expressed as $\alpha$: 
\begin{align}\label{local_enhance_form}
  e_{c ,h, w} = \begin{cases}
  \alpha  \cdot e_{c, h, w}, & p \leq h \leq p+0.1 \cdot H \\
   e_{c, h, w}, & h <p \text{ or } h >p+0.1\cdot H
  \end{cases},
\end{align}
where $ e_{c ,h, w} $ represents the value of the tensor $\mathcal{F}$ at the coordinate $c,h,w$, and $p$ is the random anchor point on the height of the feature map which is within the range $[0, 0.9 \cdot H]$. The local-enhance module performs simple linear operations on tensors with zero additional parameters. 

Traditional person re-ID systems usually use global average pooling to obtain the final feature vector, which is to reduce the loss of useful information. In order to retain the enhanced feature values effectively, we adopt extra global maximum pooling in the subsequent pooling process. Given the enhanced tensor $\mathcal{F} \in \mathbb{R}^{C\times H \times W}$, the final feature vector is calculated by:
\begin{align}
  \mathcal{V}=||A(C(\mathcal{F}))+M(C(\mathcal{F}))||,
\end{align}
where $C$ is convolution operation of layer4 of ResNet50, $A$ and $M$ respectively represent global average pooling and global maximum pooling, $||\cdot ||$ is L2-normalization and $\mathcal{V}$ is the feature vector. 
Experiments \ref{sec:ablation_study} have also demonstrated the effectiveness of the plug-and-play module. 

\section{Experiments}
All experiments were performed on a single NVIDIA Tesla P100 GPU with PyTorch 1.1. Code is available at \url{https://gitee.com/yuyetinghua/pdl/tree/center}.

\subsection{Datasets and Evaluation Metrics}
Market-1501 \cite{zheng2015scalable} and DukeMTMC-reID \cite{ristani2016performance} are widely used large-scale re-ID image datasets. We evaluate our model in two mainstream  adaptation tasks Market-to-Duke and Duke-to-Market as shown in Tabel \ref{comparison_1}.

Market-1501 \cite{zheng2015scalable} was collected on the campus of Tsinghua University, shot in the summer, and released publicly in 2015. It includes 32668 person images of 1501 identities captured by 6 cameras. Each identity is captured by at least 2 cameras, and there may be multiple images in one camera. The training set has 751 people, containing 12,936 images; the test set has 750 people, containing 19,732 images. 

DukeMTMC-reID \cite{ristani2016performance} was collected at Duke University in the winter of 2017. The DukeMTMC dataset \cite{ristani2016performance} is a large-scale multi-target multi-camera pedestrian tracking dataset. It provides a new large HD video data set recorded by 8 simultaneous cameras, with more than 7,000 single camera tracks and more than 2,700 independent characters. DukeMTMC-reID, containing 16,522 person images of 702 identities, is a person re-ID subset of the DukeMTMC dataset and provides manually marked bounding boxes.

For re-ID evaluation metrics, cumulative matching characteristics (CMC) \cite{CMC} and mean average precision (mAP) are the most commonly used evaluation indicators.  The former reflects the retrieval precision and the latter reflects the recall. We report the Rank-1, Rank-5, Rank-10, scores to represent the CMC curve.

\subsection{Implementation Details}\label{sec:details} 
We adopt ResNet50 with the last classification layer removed as our backbone network for encoder $f_\theta$. Then we add the local-enhance module between layer3 and layer4 of ResNet50 which is only enabled during training. At the first beginning, the model is initialized by pre-trained weights on ImageNet \cite{krizhevsky2012imagenet}. The batch size is set to 64, which contains 4 identities, each with 16 images on tasks Market-to-Duke and Duke-to-Market. All images are resized to regular $256\times 128$ and applied random perturbations to each image, e.g. randomly erasing, cropping, and flipping before being fed into the networks and are encoded into 2048-dimensional feature vectors by encoder $f_\theta$. And we adopt an ADAM optimizer with a weight decay of 0.0005 to train the model. The initial learning rate is set to 0.00035 and is decreased to 1/10 of its previous value every 20 epochs in a total of 50 epochs. In all experiments, only the labels of the source domain were used, and no post-processing operations, such as re-ranking, were used.

We utilize DBSCAN \cite{MartinEster1996ADA} for clustering the unlabeled target-domain data during prototype dictionary initialization and Jaccard distance \cite{Jaccard}  is used as an indicator of the algorithm to measure density. The hyperparameter $eps$ which is the max neighbor distance for DBSCAN is set to 0.6 in task Market-to-Duke and 0.5 in task Duke-to-Market and the domain density threshold $MinPts$ is set as 4. The hyperparameter $K$ which means k-reciprocal nearest neighbors is set as 30. The prototype dictionary is updated by formula \ref{momentum_update} where the hyperparameter $m$ which means momentum coefficient is set as 0.1.

\subsection{Comparison with State-of-the-Art Methods}
In this section, we compare the proposed method with state-of-the-art unsupervised cross-domain methods on the two most representative domain adaptation tasks, Market-to-Duke and Duke-to-Market in Table \ref{comparison_1}. They can be roughly divided into three categories: methods based on feature alignment, such as TJ-AIDL \cite{wang2018transferable}, methods based on GAN, such as SPGAN \cite{Deng_2018_CVPR}, and methods based on pseudo-label prediction, such as SSG \cite{fu2019self}. It can be seen that our proposed method outperforms recent works by large margins. Specifically, the mAP of our method surpasses the GAN-based method SPGAN \cite{Deng_2018_CVPR} by 49.2\% and 61.1\% on tasks Market-to-Duke and Duke-to-Market. Moreover, our method surpasses clustering-based SSG \cite{fu2019self} by a considerable margin of mAP 18.1\% and Rank-1 accuracy 10.1\% on Market-to-Duke, while 25.6\% and 13.0\% on Duke-to-Market. Meanwhile, our PDL only one encoder used is compared with the MMT \cite{ge2020mutual} method that uses the two-stream network, showing a noticeable 6.4\% and 14.9\% improvements in terms of mAP on Market-to-Duke and Duke-to-Market respectively. Compared to the state-of-the-art method GLT \cite{Zheng2021Group} which is based on label transfer, our method leads to 2.3\% mAP and 4.4\% mAP gain on Market-to-Duke and Duke-to-Market respectively. The results verify the effectiveness of our method.
\begin{table*}[!t]
  \centering
  \caption{Comparison with SOTA methods}
  \label{comparison_1}
  
\resizebox{\textwidth}{!}{
  \begin{tabular}{|c|c|c|c|c|c|c|c|c|}
  \hline
  \multicolumn{1}{|c|}{\multirow{2}{*}{Methods}} & \multicolumn{4}{c|}{Market-to-Duke} & \multicolumn{4}{c|}{Duke-to-Market} \\ \cline{2-9} 
  \multicolumn{1}{|c|}{}                                     & mAP   & rank-1  & rank-5  & rank-10 & mAP   & rank-1  & rank-5  & rank-10 \\ \hline
  TJ-AIDL \cite{wang2018transferable} (CVPR’18)              & 23.0  & 44.3    & 59.6    & 65.0    & 26.5  & 58.2    & 74.8    & 81.1    \\ 
  UCDA-CCE \cite{qi2019novel} (CVPR’19)                      & 31.0  & 47.7    & -       & -       & 30.9  & 60.4    & -       & -       \\ \hline
  SPGAN \cite{Deng_2018_CVPR} (CVPR’18)                      & 22.3  & 41.1    & 56.6    & 63.0    & 22.8  & 51.5    & 70.1    & 76.8    \\
  HHL \cite{zhong2018generali} (ECCV'2018)                   & 27.2  & 46.9    & 61.0    & 66.7    & 31.4  & 62.2    & 78.8    & 84.0    \\ 
  ECN \cite{zhong2019invariance} (CVPR’19)                   & 40.4  & 63.3    & 75.8    & 80.4    & 43.0  & 75.1    & 87.6    & 91.6    \\ 
  PDA-Net \cite{li2019cross}  (ICCV’19)                      & 45.1  & 63.2    & 77.0    & 82.5    & 47.6  & 75.2    & 86.3    & 90.2    \\ \hline
  PCB-PAST \cite{zhang2019self}  (ICCV’19)                   & 54.3  & 72.4    & -       & -       & 54.6  & 78.4    & -       & -       \\ 
  SSG \cite{fu2019self} (ICCV’19)                            & 53.4  & 73.0    & 80.6    & 83.2    & 58.3  & 80.0    & 90.0    & 92.4    \\ 
  AD-Cluster \cite{zhai2020ad} (CVPR’20)                     & 54.1  & 72.6    & -       &  -      & 68.3  & 86.7    & -       & -       \\
  MMT\cite{ge2020mutual}(ICLR’20)                            & 65.1  & 78.0    & 88.8    & 92.5    & 69.0  & 86.8    & 94.6    & 96.9    \\
  SpCL\cite{ge2020self}(NIPS’20)                             & 68.8  & 82.9    & 90.1    & 92.5    & 76.7  & 90.3    & 96.2    & 97.7    \\ 
  GLT\cite{Zheng2021Group}(CVPR’21)                          & 69.2  & 82.0    & 90.2    & 92.8    & 79.5  & 92.2    & 96.5    & 97.8    \\ \hline
  ours                   & \textbf{71.5}  & \textbf{83.1}    & \textbf{90.6}    & \textbf{92.9}    & \textbf{83.9}  & \textbf{93.0}    & \textbf{97.0}    & \textbf{98.1}    \\ \hline
  Supervised learning (BOT \cite{HaoLuo2020ASB})             & 75.8  & 86.2    & -       & -       & 85.7  & 94.1   & -        & -       \\ \hline
  \end{tabular}}
\end{table*}
\subsection{Ablation Studies}\label{sec:ablation_study}
In this section, we evaluate each component of our proposed framework PDL by conducting ablation studies on Market-to-Duke and Duke-to-Market tasks. The experimental results are shown in Table \ref{ablation_studies}. 

\textbf{Effectiveness of the category-level contrastive learning. }

The comparison of the performance of the baseline method and the method of baseline without the category-level contrastive learning are specified in Table \ref{ablation_studies}.  To verify the necessity of category-level contrastive learning, we conduct experiments by replacing instance-level comparisons with category-level comparisons. Baseline w/o cl means experiments without category-level comparisons strategy. As shown in the table, we improve the performance by 3.6\% at mAP and 3.2\% at rank-1 accuracy on Market-to-Duke and 13.8\% at mAP and 7.4\% at rank-1 accuracy on Duke-to-Market when the category-level comparisons strategy is adopted. 

 \begin{table}[!t]
   \caption{Ablation studies for our proposed framework on individual components. }
   \label{ablation_studies}
   \centering
   \begin{tabular}{|c|c|c|c|c|}
     \hline
     \multirow{2}{*}{Method}&
     \multicolumn{4}{c|}{Market-to-Duke}\\
     \cline{2-5}
                                 &mAP  &rank-1 &rank-5 &rank-10\\\hline
     baseline w/o cl             & 61.7  & 76.5    & 86.4    & 88.6 \cr\hline   
     baseline                    & 65.3  & 79.7    & 87.1    & 90.0 \cr\hline
     baseline+le                 & 68.0  & 81.3    & 89.3    & 91.6 \cr\hline
     baseline+gmp                & 70.6  & 82.3    & 90.4    & 92.3 \cr\hline
     baseline+le+gmp             & 71.5  & 83.1    & 90.6    & 92.9 \cr\hline
     \multirow{2}{*}{Method}&
     \multicolumn{4}{c|}{Duke-to-Market}\\
     \cline{2-5}
                                 &mAP  &rank-1 &rank-5 &rank-10\\\hline
     baseline w/o cl             & 66.6  & 83.8    & 93.5    & 95.5 \cr\hline
     baseline                    & 80.4  & 91.2    & 96.2    & 97.3 \cr\hline
     baseline+le                 & 83.0  & 92.3    & 96.9    & 97.7 \cr\hline
     baseline+gmp                & 81.8  & 91.3    & 96.2    & 97.3 \cr\hline
     baseline+le+gmp             & 83.9  & 93.0    & 97.0    & 98.3 \cr\hline
   \end{tabular} 
 \end{table}
 \textbf{Effectiveness of the local-enhance module. }
 We use the local-enhance module to reduce the model sensitivity of stytle to data in different fields and reduce the interference of training time domain gap. In Table \ref{ablation_studies}, Le and gmp denote to local-enhance module and additional global maximun pooling, respectively. We conduct experiments by Remove the modules added on the baseline in turn. Local-enhance module brings 2.7\% and 3.2\% improvements in terms of mAP and rank-1 accuracy on Market-to-Duke and 3.0\% and 1.1\% improvements in terms of mAP and rank-1 accuracy on Duke-to-Market, respectively. Additional global maximun pooling brings 5.3\% and 2.6\% improvements in terms of mAP and rank-1 accuracy on Market-to-Duke and 1.4\% and 0.1\% improvements in terms of mAP and rank-1 accuracy on Duke-to-Market, respectively. When both modules are added, the performance is improved by 6.2\% mAP and 3.4 rank-1 accuracy on Market-to-Duke. On Duke-to-Market, the improvements in terms of mAP and rank-1 accuracy reached 3.5\% and 1.8\% respectively. 

 We also explored the effect of the enhancement coefficients $\alpha$ on loca-enhance modules as shown in Equation \ref{local_enhance_form}. The greater the $\alpha$, the greater the change in the mean and variance of the image feature map. Because the local enhancement module is improved in both tasks, we only studied the effect of the enhancement coefficient in task Duke-to-Market, and finally chose 2 as the enhancement coefficient for the two tasks. Table \ref{enhance_coe} shows the effect of different $\alpha$ values on performance. 
\begin{table}[!t]
  \caption{Performances on Duke-to-Market with different of enhancement coefficients $\alpha$. }
   \label{enhance_coe}
   \centering
   \begin{tabular}{|c|cc|}
    \hline
    \multirow{2}{*}{$\alpha$} & \multicolumn{2}{c|}{Duke-to-Market}                \\ \cline{2-3} 
                           & \multicolumn{1}{c|}{mAP}           & rank-1        \\ \hline
    0.1                    & \multicolumn{1}{c|}{80.9}          & 91.1          \\ \hline
    0.5                    & \multicolumn{1}{c|}{81.2}          & 91.4          \\ \hline
    1                      & \multicolumn{1}{c|}{80.4}          & 91.2          \\ \hline
    1.5                    & \multicolumn{1}{c|}{81.8}          & 91.1          \\ \hline
    2                      & \multicolumn{1}{c|}{\textbf{83.9}} & \textbf{93.2} \\ \hline
    2.5                    & \multicolumn{1}{c|}{82.6}          & 92.0          \\ \hline
  \end{tabular}
  \end{table}

\section{Conclusion}
In this work, we propose the Prototype Dictionary Learning(PDL), a new unsupervised domain adaptation method for the person re-ID. The key innovation of the PDL is that we propose a new training framework which store category-level feature prototypes of the source and target domains through a novel prototype dictionary, enabling simultaneous training of source and target domain data. This further enhances the mining of labeled data while simplifying the traditional UDA methods. We propose the PDL loss, helping to replace the instance-level features with the category-level features to avoid class collision. To alleviate the amplification of inconsistency in the frequency of updating the memory dictionary during training, PDL reinitializes the memory dictionary before each epoch. We propose local-enhance module to change the value of a random part of the feature map which improves the adaptability of the model to different domain styles during training, resulting in increased domain generalization of the model without increasing parameters. We have proved the effectiveness of our method through experiments which leads to 4.4\% mAP and 2.3\% mAP gain on Duke-to-Market and Market-to-Duke respectively. 

\section*{Acknowledgment}
This work was supported by the National Natural Science Foundation of China (No. 61806206, U1610124, 61772530), the Natural Science Foundation of Jiangsu Province (No. BK20180639, BK20171192), the Six Talent Peaks Project in Jiangsu Province (No. 2015-DZXX-010).

\ifCLASSOPTIONcaptionsoff
  \newpage
\fi

\bibliographystyle{IEEEtran}

\bibliography{reference}

\begin{thebibliography}{10}
\providecommand{\url}[1]{#1}
\csname url@samestyle\endcsname
\providecommand{\newblock}{\relax}
\providecommand{\bibinfo}[2]{#2}
\providecommand{\BIBentrySTDinterwordspacing}{\spaceskip=0pt\relax}
\providecommand{\BIBentryALTinterwordstretchfactor}{4}
\providecommand{\BIBentryALTinterwordspacing}{\spaceskip=\fontdimen2\font plus
\BIBentryALTinterwordstretchfactor\fontdimen3\font minus
  \fontdimen4\font\relax}
\providecommand{\BIBforeignlanguage}[2]{{%
\expandafter\ifx\csname l@#1\endcsname\relax
\typeout{** WARNING: IEEEtran.bst: No hyphenation pattern has been}%
\typeout{** loaded for the language `#1'. Using the pattern for}%
\typeout{** the default language instead.}%
\else
\language=\csname l@#1\endcsname
\fi
#2}}
\providecommand{\BIBdecl}{\relax}
\BIBdecl

\bibitem{chang2018multi}
X.~Chang, T.~M. Hospedales, and T.~Xiang, ``Multi-level factorisation net for
  person re-identification,'' in \emph{Proceedings of the IEEE Conference on
  Computer Vision and Pattern Recognition}, 2018, pp. 2109--2118.

\bibitem{chen2018person}
D.~Chen, S.~Zhang, W.~Ouyang, J.~Yang, and Y.~Tai, ``Person search via a
  mask-guided two-stream cnn model,'' in \emph{Proceedings of the European
  Conference on Computer Vision (ECCV)}, 2018, pp. 734--750.

\bibitem{7289409}
N.~{Martinel}, C.~{Micheloni}, and G.~L. {Foresti}, ``Kernelized saliency-based
  person re-identification through multiple metric learning,'' \emph{IEEE
  Transactions on Image Processing}, vol.~24, no.~12, pp. 5645--5658, 2015.

\bibitem{8315486}
Y.~{Cho} and K.~{Yoon}, ``Pamm: Pose-aware multi-shot matching for improving
  person re-identification,'' \emph{IEEE Transactions on Image Processing},
  vol.~27, no.~8, pp. 3739--3752, 2018.

\bibitem{zheng2019joint}
Z.~Zheng, X.~Yang, Z.~Yu, L.~Zheng, Y.~Yang, and J.~Kautz, ``Joint
  discriminative and generative learning for person re-identification,'' in
  \emph{Proceedings of the IEEE conference on computer vision and pattern
  recognition}, 2019, pp. 2138--2147.

\bibitem{8693885}
L.~{Zheng}, Y.~{Huang}, H.~{Lu}, and Y.~{Yang}, ``Pose-invariant embedding for
  deep person re-identification,'' \emph{IEEE Transactions on Image
  Processing}, vol.~28, no.~9, pp. 4500--4509, 2019.

\bibitem{wang2018learning}
G.~Wang, Y.~Yuan, X.~Chen, J.~Li, and X.~Zhou, ``Learning discriminative
  features with multiple granularities for person re-identification,'' in
  \emph{Proceedings of the 26th ACM international conference on Multimedia},
  2018, pp. 274--282.

\bibitem{zhou2019omni}
K.~Zhou, Y.~Yang, A.~Cavallaro, and T.~Xiang, ``Omni-scale feature learning for
  person re-identification,'' in \emph{Proceedings of the IEEE International
  Conference on Computer Vision}, 2019, pp. 3702--3712.

\bibitem{lin2019a}
Y.~Lin, X.~Dong, L.~Zheng, Y.~Yan, and Y.~Yang, ``A bottom-up clustering
  approach to unsupervised person re-identification,'' in \emph{Proceedings of
  the AAAI Conference on Artificial Intelligence}, vol.~33, 2019, pp.
  8738--8745.

\bibitem{yu2017cross}
H.-X. Yu, A.~Wu, and W.-S. Zheng, ``Cross-view asymmetric metric learning for
  unsupervised person re-identification,'' in \emph{Proceedings of the IEEE
  international conference on computer vision}, 2017, pp. 994--1002.

\bibitem{zhai2020ad}
Y.~Zhai, S.~Lu, Q.~Ye, X.~Shan, J.~Chen, R.~Ji, and Y.~Tian, ``Ad-cluster:
  Augmented discriminative clustering for domain adaptive person
  re-identification,'' in \emph{Proceedings of the IEEE/CVF Conference on
  Computer Vision and Pattern Recognition}, 2020, pp. 9021--9030.

\bibitem{zhang2019self}
X.~Zhang, J.~Cao, C.~Shen, and M.~You, ``Self-training with progressive
  augmentation for unsupervised cross-domain person re-identification,'' in
  \emph{Proceedings of the IEEE International Conference on Computer Vision},
  2019, pp. 8222--8231.

\bibitem{fu2019self}
Y.~Fu, Y.~Wei, G.~Wang, Y.~Zhou, H.~Shi, and T.~S. Huang, ``Self-similarity
  grouping: A simple unsupervised cross domain adaptation approach for person
  re-identification,'' in \emph{Proceedings of the IEEE International
  Conference on Computer Vision}, 2019, pp. 6112--6121.

\bibitem{ge2020mutual}
Y.~Ge, D.~Chen, and H.~Li, ``Mutual mean-teaching: Pseudo label refinery for
  unsupervised domain adaptation on person re-identification,''
  \emph{International Conference on Learning Representations}, 2020.

\bibitem{ge2020self}
Y.~Ge, F.~Zhu, D.~Chen, R.~Zhao, and H.~Li, ``Self-paced contrastive learning
  with hybrid memory for domain adaptive object re-id,'' \emph{Advances in
  Neural Information Processing Systems}, vol.~33, pp. 11\,309--11\,321, 2020.

\bibitem{he2020momentum}
K.~He, H.~Fan, Y.~Wu, S.~Xie, and R.~Girshick, ``Momentum contrast for
  unsupervised visual representation learning,'' in \emph{Proceedings of the
  IEEE/CVF Conference on Computer Vision and Pattern Recognition}, 2020, pp.
  9729--9738.

\bibitem{chen2020simple}
T.~Chen, S.~Kornblith, M.~Norouzi, and G.~Hinton, ``A simple framework for
  contrastive learning of visual representations,'' in \emph{International
  conference on machine learning}.\hskip 1em plus 0.5em minus 0.4em\relax PMLR,
  2020, pp. 1597--1607.

\bibitem{saunshi2019}
N.~Saunshi, O.~Plevrakis, S.~Arora, M.~Khodak, and H.~Khandeparkar, ``A
  theoretical analysis of contrastive unsupervised representation learning,''
  in \emph{International Conference on Machine Learning}.\hskip 1em plus 0.5em
  minus 0.4em\relax PMLR, 2019, pp. 5628--5637.

\bibitem{li2020prototypical}
J.~Li, P.~Zhou, C.~Xiong, and S.~C. Hoi, ``Prototypical contrastive learning of
  unsupervised representations,'' \emph{arXiv preprint arXiv:2005.04966}, 2020.

\bibitem{huang2017arbitrary}
X.~Huang and S.~Belongie, ``Arbitrary style transfer in real-time with adaptive
  instance normalization,'' in \emph{Proceedings of the IEEE International
  Conference on Computer Vision}, 2017, pp. 1501--1510.

\bibitem{hadsell2006dimensionality}
R.~Hadsell, S.~Chopra, and Y.~LeCun, ``Dimensionality reduction by learning an
  invariant mapping,'' in \emph{2006 IEEE Computer Society Conference on
  Computer Vision and Pattern Recognition (CVPR'06)}, vol.~2.\hskip 1em plus
  0.5em minus 0.4em\relax IEEE, 2006, pp. 1735--1742.

\bibitem{varior2016gated}
R.~R. Varior, M.~Haloi, and G.~Wang, ``Gated siamese convolutional neural
  network architecture for human re-identification,'' in \emph{European
  conference on computer vision}.\hskip 1em plus 0.5em minus 0.4em\relax
  Springer, 2016, pp. 791--808.

\bibitem{hermans2017defense}
A.~Hermans, L.~Beyer, and B.~Leibe, ``In defense of the triplet loss for person
  re-identification,'' \emph{arXiv preprint arXiv:1703.07737}, 2017.

\bibitem{caron2020unsupervised}
M.~Caron, I.~Misra, J.~Mairal, P.~Goyal, P.~Bojanowski, and A.~Joulin,
  ``Unsupervised learning of visual features by contrasting cluster
  assignments,'' \emph{arXiv preprint arXiv:2006.09882}, 2020.

\bibitem{deng2018image}
W.~Deng, L.~Zheng, Q.~Ye, G.~Kang, Y.~Yang, and J.~Jiao, ``Image-image domain
  adaptation with preserved self-similarity and domain-dissimilarity for person
  re-identification,'' in \emph{Proceedings of the IEEE conference on computer
  vision and pattern recognition}, 2018, pp. 994--1003.

\bibitem{wei2018person}
L.~Wei, S.~Zhang, W.~Gao, and Q.~Tian, ``Person transfer gan to bridge domain
  gap for person re-identification,'' in \emph{Proceedings of the IEEE
  Conference on Computer Vision and Pattern Recognition}, 2018, pp. 79--88.

\bibitem{zhong2018generali}
Z.~Zhong, L.~Zheng, S.~Li, and Y.~Yang, ``Generalizing a person retrieval model
  hetero-and homogeneously,'' in \emph{Proceedings of the European Conference
  on Computer Vision (ECCV)}, 2018, pp. 172--188.

\bibitem{Aar2018Representation}
A.~van~den Oord, Y.~Li, and O.~Vinyals, ``Representation learning with
  contrastive predictive coding,'' \emph{arXiv: Learning}, 2018.

\bibitem{Cheng2019LocalAF}
C.~Zhuang, A.~Zhai, and D.~L.~K. Yamins, ``Local aggregation for unsupervised
  learning of visual embeddings,'' in \emph{International Conference on
  Computer Vision}, 2019, pp. 6002--6012.

\bibitem{ZhirongWu2018UnsupervisedFL}
Z.~Wu, Y.~Xiong, S.~X. Yu, and D.~Lin, ``Unsupervised feature learning via
  non-parametric instance discrimination,'' in \emph{Computer Vision and
  Pattern Recognition}, 2018, pp. 3733--3742.

\bibitem{MartinEster1996ADA}
M.~Ester, H.-P. Kriegel, J.~Sander, and X.~Xu, ``A density-based algorithm for
  discovering clusters in large spatial databases with noise,'' \emph{Knowledge
  Discovery and Data Mining}, 1996.

\bibitem{CPC}
A.~van~den Oord, Y.~Li, and O.~Vinyals, ``Representation learning with
  contrastive predictive coding,'' \emph{arXiv: Learning}, 2018.

\bibitem{ghiasi2018dropblock}
G.~Ghiasi, T.-Y. Lin, and Q.~V. Le, ``Dropblock: A regularization method for
  convolutional networks,'' \emph{arXiv preprint arXiv:1810.12890}, 2018.

\bibitem{zheng2015scalable}
L.~Zheng, L.~Shen, L.~Tian, S.~Wang, J.~Wang, and Q.~Tian, ``Scalable person
  re-identification: A benchmark,'' in \emph{Proceedings of the IEEE
  international conference on computer vision}, 2015, pp. 1116--1124.

\bibitem{ristani2016performance}
E.~Ristani, F.~Solera, R.~Zou, R.~Cucchiara, and C.~Tomasi, ``Performance
  measures and a data set for multi-target, multi-camera tracking,'' in
  \emph{European Conference on Computer Vision}.\hskip 1em plus 0.5em minus
  0.4em\relax Springer, 2016, pp. 17--35.

\bibitem{CMC}
D.~Gray, S.~Brennan, and H.~Tao, ``Evaluating appearance models for
  recognition, reacquisition, and tracking,'' in \emph{Proc. IEEE international
  workshop on performance evaluation for tracking and surveillance
  (PETS)}.\hskip 1em plus 0.5em minus 0.4em\relax Citeseer, 2007, pp. 1--7.

\bibitem{krizhevsky2012imagenet}
A.~Krizhevsky, I.~Sutskever, and G.~E. Hinton, ``Imagenet classification with
  deep convolutional neural networks,'' \emph{Advances in neural information
  processing systems}, vol.~25, pp. 1097--1105, 2012.

\bibitem{Jaccard}
Z.~Zhong, L.~Zheng, D.~Cao, and S.~Li, ``Re-ranking person re-identification
  with k-reciprocal encoding,'' in \emph{Computer Vision and Pattern
  Recognition}, 2017.

\bibitem{wang2018transferable}
J.~Wang, X.~Zhu, S.~Gong, and W.~Li, ``Transferable joint attribute-identity
  deep learning for unsupervised person re-identification,'' in
  \emph{Proceedings of the IEEE Conference on Computer Vision and Pattern
  Recognition}, 2018, pp. 2275--2284.

\bibitem{Deng_2018_CVPR}
W.~Deng, L.~Zheng, Q.~Ye, G.~Kang, Y.~Yang, and J.~Jiao, ``Image-image domain
  adaptation with preserved self-similarity and domain-dissimilarity for person
  re-identification,'' in \emph{Proceedings of the IEEE conference on computer
  vision and pattern recognition}, June 2018, pp. 994--1003.

\bibitem{Zheng2021Group}
K.~Zheng, W.~Liu, L.~He, T.~Mei, J.~Luo, and Z.-J. Zha, ``Group-aware label
  transfer for domain adaptive person re-identification,'' in \emph{Computer
  Vision and Pattern Recognition}, 2021.

\bibitem{qi2019novel}
L.~Qi, L.~Wang, J.~Huo, L.~Zhou, Y.~Shi, and Y.~Gao, ``A novel unsupervised
  camera-aware domain adaptation framework for person re-identification,'' in
  \emph{Proceedings of the IEEE International Conference on Computer Vision},
  2019, pp. 8080--8089.

\bibitem{zhong2019invariance}
Z.~Zhong, L.~Zheng, Z.~Luo, S.~Li, and Y.~Yang, ``Invariance matters: Exemplar
  memory for domain adaptive person re-identification,'' in \emph{Proceedings
  of the IEEE conference on computer vision and pattern recognition}, 2019, pp.
  598--607.

\bibitem{li2019cross}
Y.-J. Li, C.-S. Lin, Y.-B. Lin, and Y.-C.~F. Wang, ``Cross-dataset person
  re-identification via unsupervised pose disentanglement and adaptation,'' in
  \emph{Proceedings of the IEEE International Conference on Computer Vision},
  2019, pp. 7919--7929.

\bibitem{HaoLuo2020ASB}
H.~Luo, W.~Jiang, Y.~Gu, F.~Liu, X.~Liao, S.~Lai, and J.~Gu, ``A strong
  baseline and batch normalization neck for deep person re-identification,''
  \emph{IEEE Transactions on Multimedia}, vol.~22, pp. 2597--2609, 2020.

\end{thebibliography}

\end{document}